\documentclass{article} 

\usepackage[final]{neurips_2020}

\usepackage[utf8]{inputenc} 
\usepackage[T1]{fontenc}    
\usepackage{hyperref}       
\usepackage{url}            
\usepackage{booktabs}       
\usepackage{amsfonts}       
\usepackage{nicefrac}       
\usepackage{microtype}      
\usepackage{float}
\usepackage{graphicx}
\usepackage{amsmath}
\usepackage{natbib}
\usepackage{apalike}
\usepackage{verbatim} 
\usepackage{multirow}
\usepackage{commath}

\title{A Biologically-Inspired Dual Stream World Model}

\author{Arthur Juliani \\
Department of Psychology \\
University of Oregon \\
\texttt{ajuliani@uoregon.edu} \\
\And
Margaret Sereno \\
Department of Psychology \\
University of Oregon\\
\texttt{msereno@uoregon.edu} \\
}

\begin{document}

\maketitle

\begin{abstract}
The medial temporal lobe (MTL), a brain region containing the hippocampus and nearby areas, is hypothesized to be an experience-construction system in mammals, supporting both recall and imagination of temporally-extended sequences of events. Such capabilities are also core to many recently proposed ``world models" in the field of AI research. Taking inspiration from this connection, we propose a novel variant, the Dual Stream World Model (DSWM), which learns from high-dimensional observations and dissociates them into context and content streams. DSWM can reliably generate imagined trajectories in novel 2D environments after only a single exposure, outperforming a standard world model. DSWM also learns latent representations which bear a strong resemblance to place cells found in the hippocampus. We show that this representation is useful as a reinforcement learning basis function, and that the generative model can be used to aid the policy learning process using Dyna-like updates.
\end{abstract}

\section{Introduction}

Humans are able to recall and imagine long, temporally extended sequences of events. This capability has been referred to as the brain's `construction system,' because of the way in which these coherent sequences are constructed during memory recall, imagination, planning, and when dreaming \citep{hassabis2009construction}. The capacity to represent coherent temporal sequences of events is tied closely to the ability to skillfully navigate the world around us, due to the existence of what has been referred to as a cognitive map of space in mammals \citep{tolman1948cognitive}. Both abilities have been localized to hippocampus and surrounding structures, collectively referred to as the medial temporal lobe (MTL) \citep{tulving1998episodic,o1978hippocampus,morris1982place}. 

The spatial context represented by the cognitive map in the hippocampus has been proposed to provide an index for the experiential content of the memory itself, which is stored elsewhere in the cortex \citep{teyler1986hippocampal}. These index representations themselves spontaneously activate in coherent sequences which mirror those of animals during actual experience \citep{foster2017replay}. It has been hypothesized that these capabilities underpin numerous cognitive abilities, from planning to memory consolidation \citep{pezzulo2017internally}. There have been a wealth of proposed theoretical models for the MTL \citep{mcnaughton1991dead,hasselmo2009model,schapiro2017complementary,whittington2019tolman} (See \citep{behrens2018cognitive} for a recent review). 

In parallel, within the field of AI research, generative temporal models, often referred to as `world models' \citep{ha2018world} have been developed which are able to emulate the ability of the medial temporal lobe to generate temporally extended sequences of experience. Often these models are used within the context of model-based Reinforcement Learning, where the outcomes of simulated events in the model are used to either learn a value function, or improve a policy \citep{sutton2018reinforcement,hafner2018learning}. 

Recently proposed world models such as Generative Temporal Model with Memory (GTM-M) and Memory Based Predictor (MBP) have incorporated differentiable memory stores \citep{gemici2017generative,wayne2018unsupervised}. Such memory systems allow for adaptation to changes in the environment within a given episode, and can be seen as relating directly to the function of the hippocampus in the MTL \citep{wayne2018unsupervised}. Rather than having a single latent variable represent a given state, another class of models, including the Generative Temporal Model with Spatial Memory (GTM-SM) and the Tolman Eichenbaum Machine (TEM), split the representation into context and content variables \citep{fraccaro2018generative,whittington2019tolman}. This enables the re-use of structural knowledge when faced with novel content within an environment. 

Taking inspiration from the construction hypothesis, as well as the recent innovations described above, we propose a novel method which utilizes separate context and content streams, a differentiable memory store, and a forward model over context variables. We refer to this model as a Dual Stream World Model (DSWM), and demonstrate that it outperforms a single stream world model on a series of generative modeling tasks in environments with shared structure but novel content. In addition, it also learns a latent representation which bears a strong resemblance to that of place cells. We demonstrate that this learned representation serves as a useful basis function for downstream reinforcement learning tasks. Furthermore, by utilizing the generative model to perform additional offline learning using the Dyna algorithm \citep{sutton1991dyna,russek2017predictive}, agents using this state space are able to learn to solve navigation tasks in only a few exposures to the environment.

\section{Dual-Stream World Model}

\begin{figure}[h!]
\centering
\includegraphics[width=0.6\textwidth]{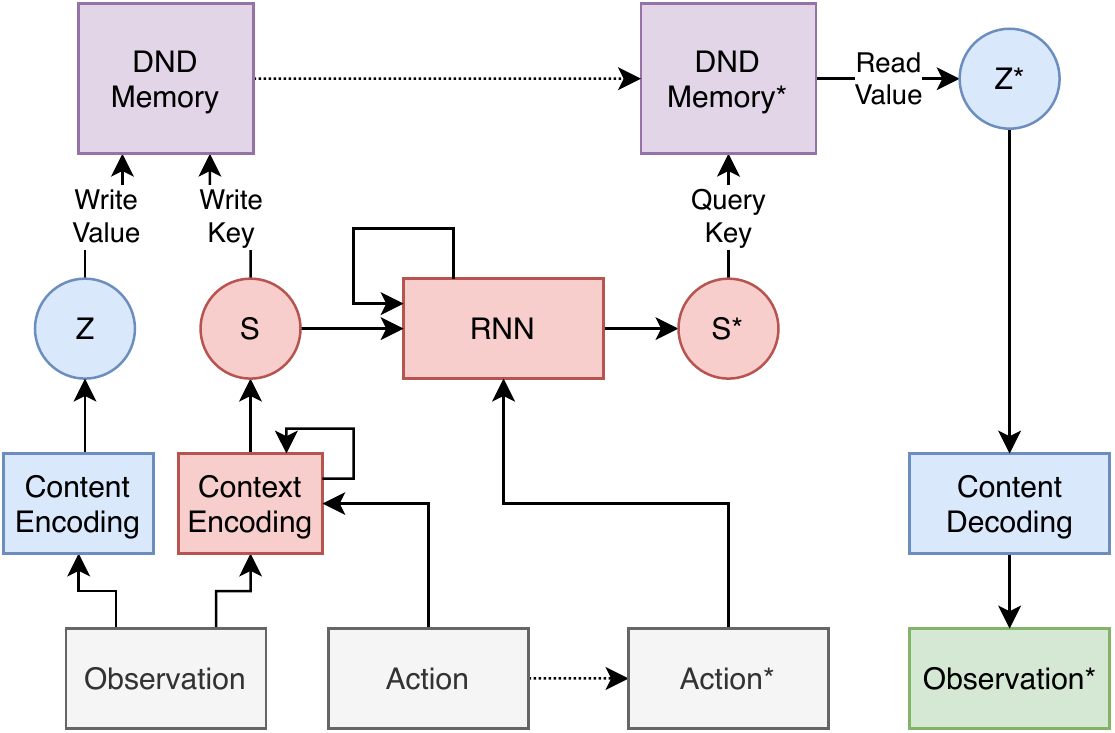}
\caption{Dual Stream World Model diagram. Blue represents content information. Red represents context information. Purple represents joint content and context information. White represents model inputs. Green represents model outputs. Nodes marked with a $*$ indicate information at the next time step of the simulation.}
\label{dswmModelDiagram}
\end{figure}

The DSWM consists of four main components. A content auto-encoder, a context encoder, a forward model (RNN), and a differentiable neural dictionary (DND). Specifically, we utilize a variational encoder with a gumbel-softmax distribution for both the context and content components \citep{kingma2013auto,jang2016categorical}. We implement the forward model using a gated recurrent unit (GRU) \citep{chung2014empirical}, and use as input both the latent context state $s$ as well as the current action $a$. The DND is similar to that introduced by \citet{pritzel2017neural} and uses the latent context state $s$ as keys, and the latent content state $z$ as values. The lookup process (DND Memory*) uses cosine similarity between a query key ($s*$) and the stored keys to determine a similarity score. The top 5 stored values are then weighted by their similarity scores using a softmax function to derive the retrieved $z$ ($z*$).

This process can be seen as roughly mapping onto the lateral entorhinal cortex (content encoding) \citep{deshmukh2011representation}, medial entorhinal cortex (context encoding) \citep{hafting2005microstructure}, and hippocampus (differentiable look-up and forward model) \citep{hassabis2009construction}. The model can be seen as an instantiation of the memory indexing theory, whereby the context variable is used to index the content variable, which itself is an abstracted representation of a high-dimensional observation, which can be thought of as a cortical state \citep{teyler1986hippocampal}.

The following series of steps take place in a given time-step. First a new observation is observed from the environment, and encoded as the latent content variable $z_t$. In parallel, the observation $o_t$ and the previous action $a_{t-1}$ are used to encode the latent context $s_t$ variable. The inferred context variable $s_t$ and content variable $z_t$ are then stored together as a key-value pair in the DND $M_t$. The forward model is then unrolled using both the next action $a_t$ the agent takes, and the current inferred context variable $s_t$ to produce a new context variable $s_{t+1}$ that is used to query the memory to read a new content variable $z_{t+1}$, which is decoded into a predicted observation $o_{t+1}$. This process is described in Figure \ref{dswmModelDiagram}. Specific details of the network model architecture are presented in \ref{modelFlow}.

The DSWM is trained to minimize four objectives. Observation prediction: mean squared error between actual and predicted observations $L_{Obs} = \frac{1}{n} \sum_{n=1}^{N}{|o^q_t - o^p_t|^2}$. Spatial context prediction: mean squared error between true and predicted position $L_{Pos} = \frac{1}{n} \sum_{n=1}^{N}{|pos^q_t - pos^p_t|^2}$. Sequence coherence: Kullback–Leibler (KL) divergence between inferred and generated context variables $L_{S} = D_{KL}(p(s_t | o, s_{t-1}) || q(s_{t+1} | s_t, a_t))$. Latent variable regularization: the negative entropy of the context and content variable distributions, which acts as a regularization term. 

We use a form of supervision to train the context variable $s$, based on agent position. We note however that other fully unsupervised loss functions are possible in cases where the environment is not inherently spatial, such as retrieval error during the look-up process.

\section{Evaluation Methods}

\subsection{Generative Modeling Methods}

\begin{figure}[h!]
\centering
\includegraphics[width=0.75\textwidth]{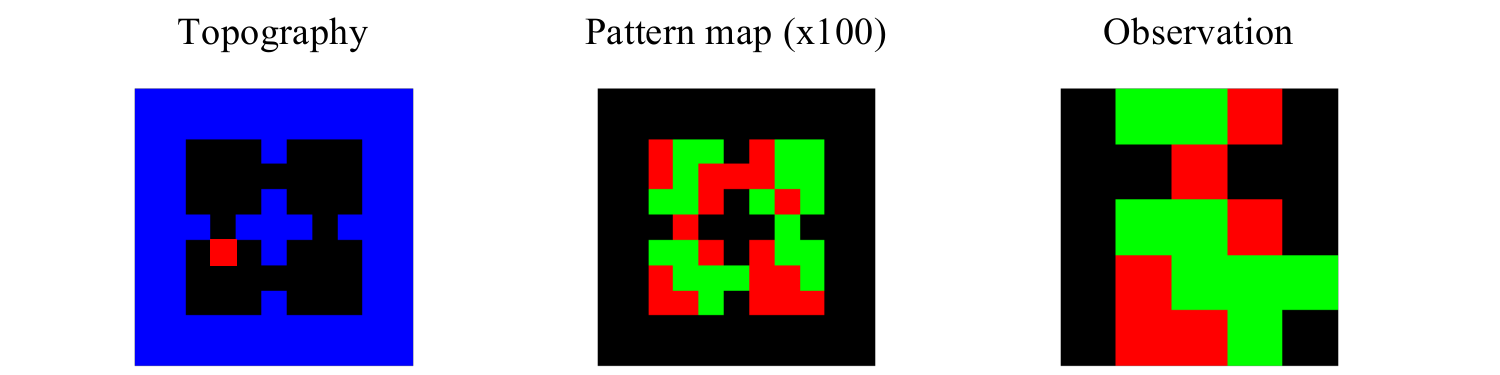}
\caption{Variable content environment. Left: example environment topography. Blue corresponds to walls. Red corresponds to agent position. Middle: Randomly generated pattern image used to derive observations based on agent location. Right: Agent observations provide a 5x5 window around the agent position.}
\label{alloVariable}
\end{figure}

In order to examine the capabilities of the DSWM to predict coherent trajectories of observations, we use a set of environments with a complex topographic structure, partially visible observations, and variability in appearance. Each environment consists of a 2D gridworld, from which the agent can move in the four cardinal directions, but cannot move through walls. Each environment is composed of $11x11$ square units. We use images drawn from a sliding window over a larger visual pattern map juxtaposed on the environment. See Figure \ref{alloVariable} for an example of these environment topographies, the pattern maps, and the derived observations.

Each pattern map is generated by randomly selecting a green or red pixel to be placed in each unit of the environment that does not contain a wall. The agent is provided with observations which consist of a $5x5$ window around its current location, which displays the content of the pattern map as well as the location of any walls within the environment. We use environments with four different topographies. These consist of an open area $Open Maze$, an environment with four connected rooms $Rooms Maze$, an environment with a symmetrical obstacle in the middle $Ring Maze$, and an environment with four symmetrical obstacles $Hallway Maze$ (See also Figure \ref{alloVariableAll} for examples of all four topographies). For each of these topographies, we generate 100 different pattern maps to provide a variety of different objects for the agent to observe. 

The datasets used to train each model were collected by running a semi-random behavioral policy for 1000 episodes of 50 steps each. In this case, we create four different datasets, one for each unique topography, and randomly select one of 100 pattern maps to use for each episode.

We compare the proposed DSWM to a single stream world model implementation with similar latent distribution types and capacity \citep{ha2018world}. See Figure \ref{worldModelDiagram} in the Appendix for a diagram of this model. Note that during training we reset the DND of the DSWM between each episode, so that stored memories do not carry over. See Table \ref{hyperparametersTable} for the complete hyperparameters used to train both the WORLD and DSWM models.

\subsection{Reinforcement Learning Methods}

\begin{figure}[h!]
\centering
\includegraphics[width=0.65\textwidth]{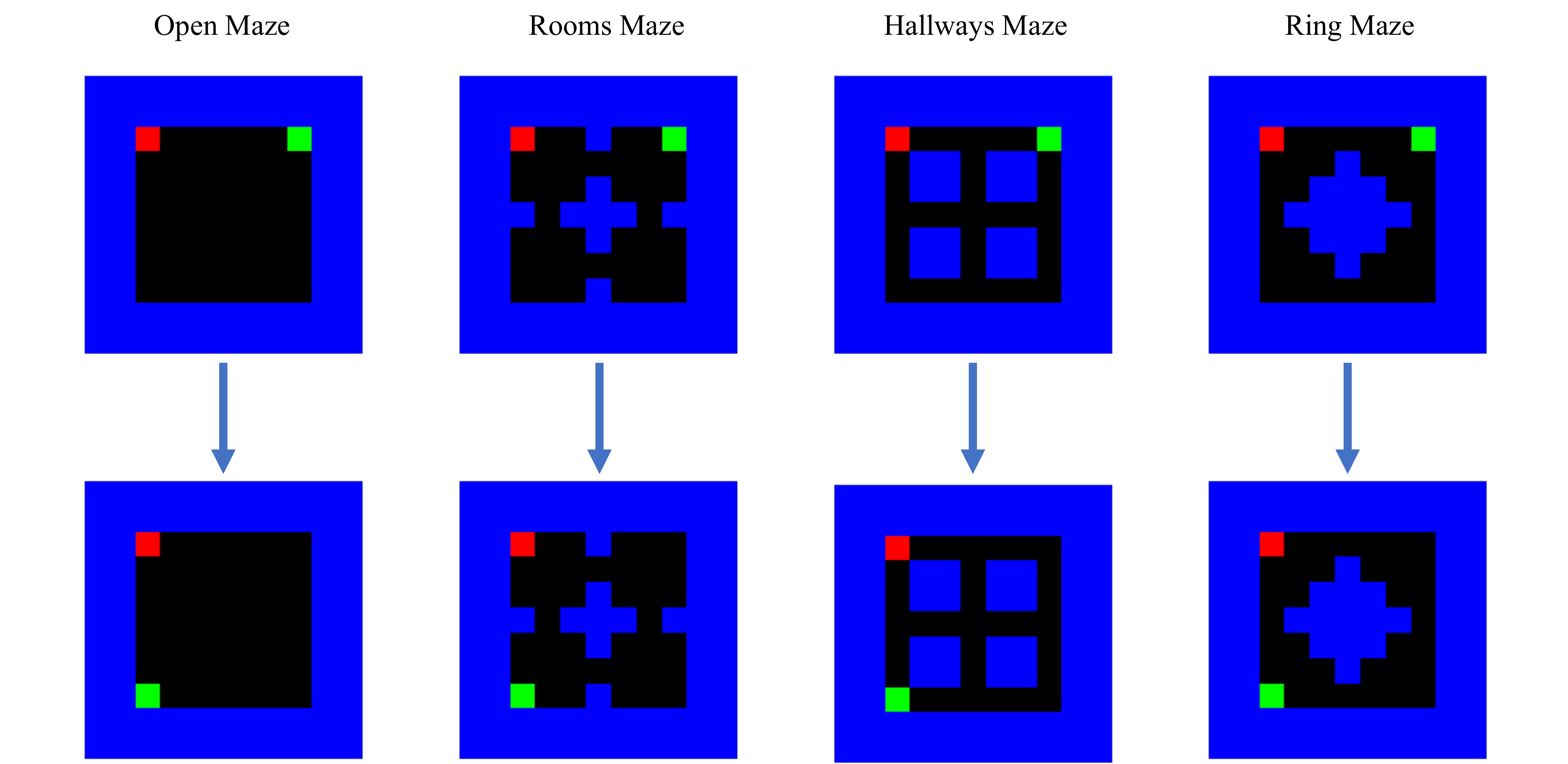}
\caption{Four different environment topographies, each showing the initial goal location for the first 50 episodes (top) and the second goal location for the following 50 episodes (bottom).}
\label{2DGoalChange}
\end{figure}

We employ a goal-directed navigation task which involves the agent searching for a hidden goal in one of the states of the environment. The reward function consists of a $+1$ reward when the agent reaches the goal state, and $0$ reward in all other states. The agent begins each episode in the same location. Halfway through a given training session, in this case, 50 episodes into training, the location of the goal changes to a new position. We use the same set of goal locations for all topographies in order to allow for the consistent comparison between results. See Figure \ref{2DGoalChange} for a visual representation of the goal locations before and after the change for each environment topography. 

We train agent policies using a modification of the successor feature algorithm \citep{barreto2017successor}, chosen for its connection to biological representations within the hippocampus \cite{stachenfeld2017hippocampus}, as well as its ability to enable rapid adaptation to changes in goal, similar to what is found in animals. In order to address probabilistic state spaces such as those induced by using a distributional representation for $s$, we replace the dot product between the successor features $\psi$ and the reward function $w$ with a cosine similarity function to derive the $Q$ and $V$ functions. This enables the use of a wider class of representations, including those which are probabilistic over state occupancy. For additional details, see Section \ref{SSA}.

Each trained agent uses one of three different basis functions: the $z$ distribution from a world model (WORLD), the $s$ distribution from the DSWM, and a pre-computed onehot encoding. We also include an additional variant where we augment the DSWM state space agent with an additional offline learning procedure based on the Dyna algorithm \citep{sutton1991dyna}. This offline algorithm uses the forward model of the DSWM to generate trajectories of $s$ states.

Agents are trained for 100 episodes each, with a maximum of 100 steps per episode using an environment from the test set of pattern maps. We reset the DND of the DSWM model between each episode. Each training session is repeated with five separate agent initialization seeds in order to better understand learning dynamics. See Table \ref{hyperparametersTableRL} for all hyperparameters used in these experiments.

\section{Experimental Results}

\subsubsection{Generative Modeling Results}

We first compare the prediction accuracy of the models' auto-regressive rollouts in a novel environment. We use a separate set of five held-out pattern maps to create five novel environments for each of the four different topographies to evaluate the models. We collect predictions based on first allowing the agent to run for 30 time-steps within an environment, and then auto-regressively predict the next 20 observations. 

\begin{figure}[h!]
\centering
\includegraphics[width=0.9\textwidth]{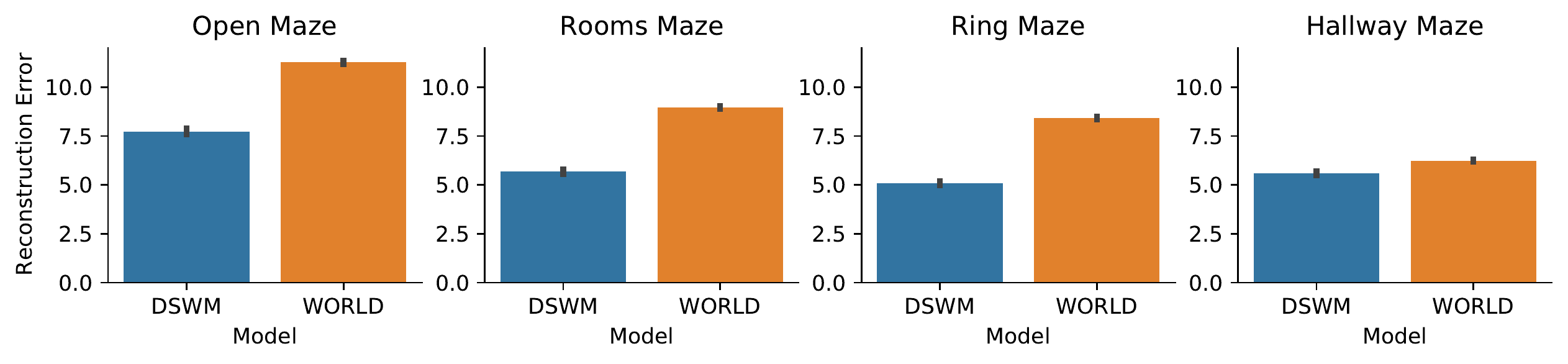}
\caption{MSE of observation predictions from rollouts of both models in four different environment topographies. Error bars represent standard error. In all environments, DSWM is able to significantly better predict trajectories of future observations than the WORLD model.}
\label{DSWM2DBarGraphs}
\end{figure}

We find that for all tested environments the DSWM is able to more accurately predict sequences of observations in these novel environments which were not part of the dataset used for training (DSWM $Mean = 6.025, Std = 6.573$, WORLD $Mean = 8.752, Std = 4.594$, $p < 0.001$). See Figure \ref{DSWM2DBarGraphs} for the individual losses within each environment. These results suggest that DSWM does indeed have additional generalization ability compared to the WORLD model.

\begin{figure}[h!]
\centering
\includegraphics[width=1.0\textwidth]{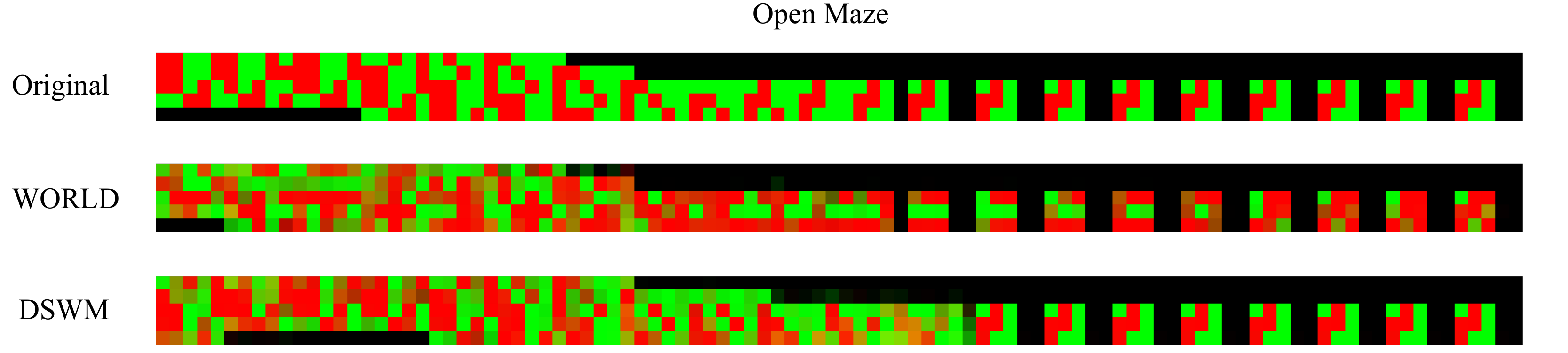}
\caption{Examples of reconstructed observations from rollouts of both World and DSWM models in the ``Open Maze." Environment uses pattern map reserved for testing, and not seen during training. DSWM is able to better predict the true trajectory of future observations within the novel environment.}
\label{DSWM2DRollouts}
\end{figure}

We can also inspect qualitatively the predictions produced by each model. Example auto-regressive rollouts from the two models are presented in Figure \ref{DSWM2DRollouts} (Rollout example from all environment topography variations are presented in Figure \ref{DSWM2DRolloutsFull}. We can see that while both models are reasonably accurate at predicting the structure of the environment, the WORLD model fails to predict the correct content in novel environments, whereas the DSWM is able to predict both the content and structure. As such, this provides evidence that the DSWM is able to adapt to an environment's novel visual content as long as it retains a familiar topographical structure. 

We next examined the learned latent representations within the DSWM, asking whether the learned representation of the $s$ latent space reflects place-like firing properties. Given the loss function which induces a representation from which the agent position can be decoded, we might expect that such a representation would arise. This is not guaranteed however, since the observations being encoded into $s$ contain both spatial and non-spatial information, and in some cases the non-spatial information dominates the observation.

\begin{figure}[h!]
\centering
\includegraphics[width=1.0\textwidth]{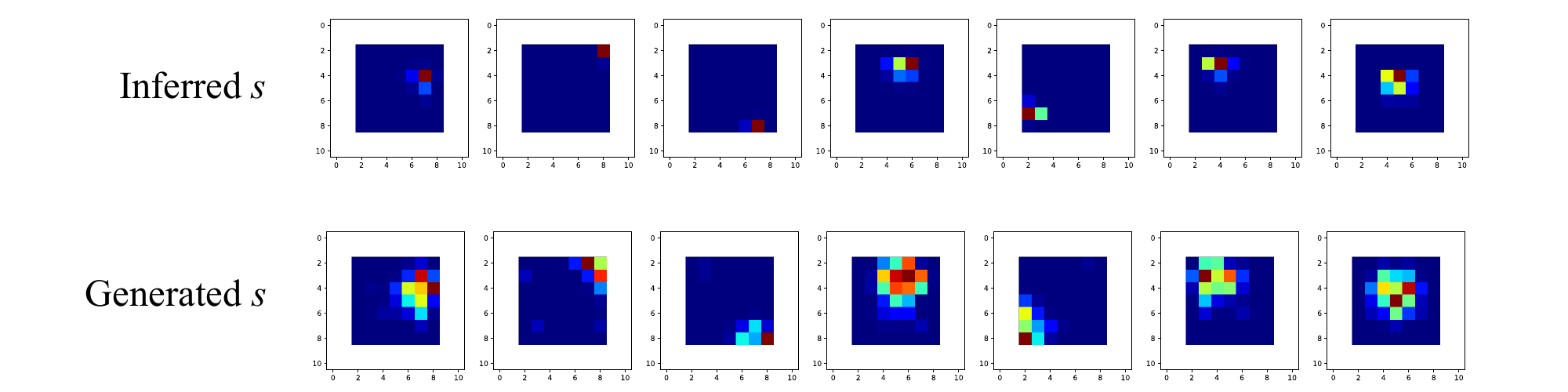}
\caption{Examples of activations of first fourteen units of inferred and generated $s$ from DSWM model in the ``Open Maze" environment topography.}
\label{DSWMPlaceCells}
\end{figure}

To answer this question, we qualitatively examine the learned representations of $s$ mapped onto the environment topography. We find that the representations can be best described as indeed being place-like in their firing affinities. See Figure \ref{DSWMPlaceCells} for examples. In particular, we find that the inferred $s_t$ units are highly spatially local, whereas the $s_{t+1}$ units generated by the forward model have wider spatial selectivity. We can draw a hypothetical connection to the CA3 and CA1 regions of the hippocampus, which are hypothesized to be involved in inference and generation, respectively \citep{teyler2007hippocampal}. 

\subsubsection{Reinforcement Learning Results}

We present the results of the goal-driven navigation experiments in Figure \ref{dswmRLResultsTable} (Additionally, see Figure \ref{dswmRLResults} for learning curves). This contains the mean and median steps-to-goal of the final 20 episodes of training for each agent. We find that for all four environments, the state space derived from the DSWM model latent space $s$ is able to match or outperform both the state space derived from the WORLD model latent space $s$ as well as the one-hot state space encoding. 

\begin{figure}[h!]
    \centering
    \begin{center}
    \begin{tabular}{ |c|c|c|c|c|c|c| } 
    \hline
    Topography & Optimal & Statistic & WORLD & DSWM & DSWM+DYNA & ONEHOT \\
    \hline
    \hline
    \multirow{2}{4em}{Open} & \multirow{2}{4em}{5} & Mean & 32.1 & 5.81 & 5.0 & 7.76 \\ 
    & & Median & 7.45 & 5.0 & 5.0 & 7.1 \\ 
    \hline
    \multirow{2}{4em}{Rooms} & \multirow{2}{4em}{7} & Mean & 99.0 & 23.93 & 7.04 & 8.64 \\ 
    & & Median & 99.0 & 7.6 & 7.0 & 7.55 \\ 
    \hline
    \multirow{2}{4em}{Ring} & \multirow{2}{4em}{5} & Mean & 99.0 & 23.8 & 5.0 & 5.0 \\ 
    & & Median & 99.0 & 5.0 & 5.0 & 5.0 \\ 
    \hline
    \multirow{2}{4em}{Hallway} & \multirow{2}{4em}{5} & Mean & 79.22 & 5.0 & 5.0 & 5.0 \\ 
    & & Median & 99.0 & 5.0 & 5.0 & 5.0 \\ 
    \hline
    \end{tabular}
    \end{center}
    \caption{Statistics from final 20 episodes of each training session for goal-directed agents. DSWM+DYNA results in most consistent learning, with near optimal performance in all four topographies.}
    \label{dswmRLResultsTable}
\end{figure}

We furthermore find that in all environment topographies, the addition of the Dyna algorithm improves the performance of the DSWM state space-based agents, and results in optimal performance (shortest route) for three out of the four environments, with the ``Rooms Maze" performance being nearly optimal. We can interpret these results as a clear sign that the learned latent space in the DSWM model is both useful for predicting trajectories of experience in novel environments and in supporting goal-directed navigation in novel environments. Additionally the DSWM+DYNA model performing best suggests that the DSWM has learned a coherent model of the dynamics of the environment which are able to abstract away the specific content of the environment. 

\section{Conclusion}

In this work, we introduced the Dual Stream World Model, a novel generative temporal model which takes inspiration from the `construction system' of the medial temporal lobe. We analyzed this novel model with respect to the coherent generation of trajectories of experience, and found that it is better able to predict future trajectories of experience than a standard world model. Furthermore, we found that the latent context representation within the model bears a strong resemblance to hippocampal place cells, and validated this latent space by demonstrating its usefulness in supporting goal-directed navigation. 

The DSWM can be seen as one of a class of recent generative temporal models, such as the Model-Based Predictor (MBP) \citep{wayne2018unsupervised}, the Generative Temporal Model with Spatial Memory (GTM-SM) \citep{fraccaro2018generative}, and the Tolman-Eichenbaum Machine (TEM) \citep{whittington2019tolman}. We believe that the DSWM meaningful addition to this growing ensemble of memory-based models of hippocampal learning. It has clearly demonstrated properties of adaptability to changes in environmental content, both in terms of generating coherent trajectories of experience, and in supporting goal-directed navigation, both key properties of a flexible cognitive map.

\pagebreak

\bibliographystyle{apalike}
\bibliography{reference.bib}

\begin{thebibliography}{}

\bibitem[Barreto et~al., 2017]{barreto2017successor}
Barreto, A., Dabney, W., Munos, R., Hunt, J.~J., Schaul, T., van Hasselt,
  H.~P., and Silver, D. (2017).
\newblock Successor features for transfer in reinforcement learning.
\newblock In {\em Advances in neural information processing systems}, pages
  4055--4065.

\bibitem[Behrens et~al., 2018]{behrens2018cognitive}
Behrens, T.~E., Muller, T.~H., Whittington, J.~C., Mark, S., Baram, A.~B.,
  Stachenfeld, K.~L., and Kurth-Nelson, Z. (2018).
\newblock What is a cognitive map? organizing knowledge for flexible behavior.
\newblock {\em Neuron}, 100(2):490--509.

\bibitem[Chung et~al., 2014]{chung2014empirical}
Chung, J., Gulcehre, C., Cho, K., and Bengio, Y. (2014).
\newblock Empirical evaluation of gated recurrent neural networks on sequence
  modeling.
\newblock {\em arXiv preprint arXiv:1412.3555}.

\bibitem[Deshmukh and Knierim, 2011]{deshmukh2011representation}
Deshmukh, S.~S. and Knierim, J.~J. (2011).
\newblock Representation of non-spatial and spatial information in the lateral
  entorhinal cortex.
\newblock {\em Frontiers in behavioral neuroscience}, 5:69.

\bibitem[Foster, 2017]{foster2017replay}
Foster, D. (2017).
\newblock Replay comes of age.
\newblock {\em Annual review of neuroscience}, 40:581--602.

\bibitem[Fraccaro et~al., 2018]{fraccaro2018generative}
Fraccaro, M., Rezende, D.~J., Zwols, Y., Pritzel, A., Eslami, S., and Viola, F.
  (2018).
\newblock Generative temporal models with spatial memory for partially observed
  environments.
\newblock {\em arXiv preprint arXiv:1804.09401}.

\bibitem[Gemici et~al., 2017]{gemici2017generative}
Gemici, M., Hung, C.-C., Santoro, A., Wayne, G., Mohamed, S., Rezende, D.~J.,
  Amos, D., and Lillicrap, T. (2017).
\newblock Generative temporal models with memory.
\newblock {\em arXiv preprint arXiv:1702.04649}.

\bibitem[Ha and Schmidhuber, 2018]{ha2018world}
Ha, D. and Schmidhuber, J. (2018).
\newblock World models.
\newblock {\em arXiv preprint arXiv:1803.10122}.

\bibitem[Hafner et~al., 2018]{hafner2018learning}
Hafner, D., Lillicrap, T., Fischer, I., Villegas, R., Ha, D., Lee, H., and
  Davidson, J. (2018).
\newblock Learning latent dynamics for planning from pixels.
\newblock {\em arXiv preprint arXiv:1811.04551}.

\bibitem[Hafting et~al., 2005]{hafting2005microstructure}
Hafting, T., Fyhn, M., Molden, S., Moser, M.-B., and Moser, E.~I. (2005).
\newblock Microstructure of a spatial map in the entorhinal cortex.
\newblock {\em Nature}, 436(7052):801.

\bibitem[Hassabis and Maguire, 2009]{hassabis2009construction}
Hassabis, D. and Maguire, E.~A. (2009).
\newblock The construction system of the brain.
\newblock {\em Philosophical Transactions of the Royal Society B: Biological
  Sciences}, 364(1521):1263--1271.

\bibitem[Hasselmo, 2009]{hasselmo2009model}
Hasselmo, M.~E. (2009).
\newblock A model of episodic memory: mental time travel along encoded
  trajectories using grid cells.
\newblock {\em Neurobiology of learning and memory}, 92(4):559--573.

\bibitem[Jang et~al., 2016]{jang2016categorical}
Jang, E., Gu, S., and Poole, B. (2016).
\newblock Categorical reparameterization with gumbel-softmax.
\newblock {\em arXiv preprint arXiv:1611.01144}.

\bibitem[Kingma and Welling, 2013]{kingma2013auto}
Kingma, D.~P. and Welling, M. (2013).
\newblock Auto-encoding variational bayes.
\newblock {\em arXiv preprint arXiv:1312.6114}.

\bibitem[McNaughton et~al., 1991]{mcnaughton1991dead}
McNaughton, B.~L., Chen, L., and Markus, E. (1991).
\newblock “dead reckoning,” landmark learning, and the sense of direction:
  a neurophysiological and computational hypothesis.
\newblock {\em Journal of Cognitive Neuroscience}, 3(2):190--202.

\bibitem[Morris et~al., 1982]{morris1982place}
Morris, R.~G., Garrud, P., Rawlins, J.~a., and O'Keefe, J. (1982).
\newblock Place navigation impaired in rats with hippocampal lesions.
\newblock {\em Nature}, 297(5868):681.

\bibitem[O'keefe and Nadel, 1978]{o1978hippocampus}
O'keefe, J. and Nadel, L. (1978).
\newblock {\em The hippocampus as a cognitive map}.
\newblock Oxford: Clarendon Press.

\bibitem[Paszke et~al., 2019]{paszke2019pytorch}
Paszke, A., Gross, S., Massa, F., Lerer, A., Bradbury, J., Chanan, G., Killeen,
  T., Lin, Z., Gimelshein, N., Antiga, L., et~al. (2019).
\newblock Pytorch: An imperative style, high-performance deep learning library.
\newblock In {\em Advances in neural information processing systems}, pages
  8026--8037.

\bibitem[Pezzulo et~al., 2017]{pezzulo2017internally}
Pezzulo, G., Kemere, C., and Van Der~Meer, M.~A. (2017).
\newblock Internally generated hippocampal sequences as a vantage point to
  probe future-oriented cognition.
\newblock {\em Annals of the New York Academy of Sciences}, 1396(1):144--165.

\bibitem[Pritzel et~al., 2017]{pritzel2017neural}
Pritzel, A., Uria, B., Srinivasan, S., Badia, A.~P., Vinyals, O., Hassabis, D.,
  Wierstra, D., and Blundell, C. (2017).
\newblock Neural episodic control.
\newblock In {\em Proceedings of the 34th International Conference on Machine
  Learning-Volume 70}, pages 2827--2836. JMLR. org.

\bibitem[Ramachandran et~al., 2017]{ramachandran2017searching}
Ramachandran, P., Zoph, B., and Le, Q.~V. (2017).
\newblock Searching for activation functions.
\newblock {\em arXiv preprint arXiv:1710.05941}.

\bibitem[Russek et~al., 2017]{russek2017predictive}
Russek, E.~M., Momennejad, I., Botvinick, M.~M., Gershman, S.~J., and Daw,
  N.~D. (2017).
\newblock Predictive representations can link model-based reinforcement
  learning to model-free mechanisms.
\newblock {\em PLoS computational biology}, 13(9):e1005768.

\bibitem[Schapiro et~al., 2017]{schapiro2017complementary}
Schapiro, A.~C., Turk-Browne, N.~B., Botvinick, M.~M., and Norman, K.~A.
  (2017).
\newblock Complementary learning systems within the hippocampus: a neural
  network modelling approach to reconciling episodic memory with statistical
  learning.
\newblock {\em Philosophical Transactions of the Royal Society B: Biological
  Sciences}, 372(1711):20160049.

\bibitem[Stachenfeld et~al., 2017]{stachenfeld2017hippocampus}
Stachenfeld, K.~L., Botvinick, M.~M., and Gershman, S.~J. (2017).
\newblock The hippocampus as a predictive map.
\newblock {\em Nature neuroscience}, 20(11):1643.

\bibitem[Sutton, 1991]{sutton1991dyna}
Sutton, R.~S. (1991).
\newblock Dyna, an integrated architecture for learning, planning, and
  reacting.
\newblock {\em ACM SIGART Bulletin}, 2(4):160--163.

\bibitem[Sutton and Barto, 2018]{sutton2018reinforcement}
Sutton, R.~S. and Barto, A.~G. (2018).
\newblock {\em Reinforcement learning: An introduction}.
\newblock MIT press.

\bibitem[Teyler and DiScenna, 1986]{teyler1986hippocampal}
Teyler, T.~J. and DiScenna, P. (1986).
\newblock The hippocampal memory indexing theory.
\newblock {\em Behavioral neuroscience}, 100(2):147.

\bibitem[Teyler and Rudy, 2007]{teyler2007hippocampal}
Teyler, T.~J. and Rudy, J.~W. (2007).
\newblock The hippocampal indexing theory and episodic memory: updating the
  index.
\newblock {\em Hippocampus}, 17(12):1158--1169.

\bibitem[Tolman, 1948]{tolman1948cognitive}
Tolman, E.~C. (1948).
\newblock Cognitive maps in rats and men.
\newblock {\em Psychological review}, 55(4):189.

\bibitem[Tulving and Markowitsch, 1998]{tulving1998episodic}
Tulving, E. and Markowitsch, H.~J. (1998).
\newblock Episodic and declarative memory: role of the hippocampus.
\newblock {\em Hippocampus}, 8(3):198--204.

\bibitem[Wayne et~al., 2018]{wayne2018unsupervised}
Wayne, G., Hung, C.-C., Amos, D., Mirza, M., Ahuja, A., Grabska-Barwinska, A.,
  Rae, J., Mirowski, P., Leibo, J.~Z., Santoro, A., et~al. (2018).
\newblock Unsupervised predictive memory in a goal-directed agent.
\newblock {\em arXiv preprint arXiv:1803.10760}.

\bibitem[Whittington et~al., 2019]{whittington2019tolman}
Whittington, J.~C., Muller, T.~H., Mark, S., Chen, G., Barry, C., Burgess, N.,
  and Behrens, T.~E. (2019).
\newblock The tolman-eichenbaum machine: Unifying space and relational memory
  through generalisation in the hippocampal formation.
\newblock {\em bioRxiv}, page 770495.

\end{thebibliography}

\clearpage

\appendix

\section{Supplemental Material}

\subsection{Dual Stream World Model Implementation Details}\label{modelFlow}

At each time-step of simulation, the Dual Stream World model operates in two phases, an inference and a generation phase. These phases are governed by the following equations.

Inference phase:
\begin{align}
    z_t &\sim p_{enc}(z_t | o_t) \\
    s_t &\sim p_{enc}(s_t | o_t) \\
    M_t &= f_{write}(M_{t-1}, s_t, z_t) \\
    h_t &= f_{forward}(s_t, a_t, h_{t-1})
\end{align}

Generation phase:
\begin{align}
    s_{t+1} &\sim q_{forward}(s_{t+1} | s_t, a_t, h_t) \\
    z_{t+1} &\sim q_{read}(z_{t+1} | M_t, s_{t+1}) \\
    o_{t+1} &= f_{decode}(z_{t+1})
\end{align}

This flow, as well as that of the WORLD model used as a comparison baseline are concretely implemented as a fully-differentiable neural networks written using the PyTorch framework \citep{paszke2019pytorch}. In DSWM, $p_{enc}(z_t | o_t)$ and $p_{enc}(s_t | o_t)$ are implemented as three layer multi-layer perceptrons with 256 hidden units each, using the $swish$ activation function after each layer \citep{ramachandran2017searching}, except for the final layer, which consists of a gumbel-softmax distribution \citep{jang2016categorical}. $f_{forward}(s_t, a_t, h_{t-1})$ and $q_{forward}(s_{t+1} | s_t, a_t, h_t)$ are implemented as a gated recurrent unit (GRU) with a hidden layer size of 256 units \citep{chung2014empirical}. $f_{write}(M_{t-1}, s_t, z_t)$ and $q_{read}(z_{t+1} | M_t, s_{t+1})$ are implemented as a differentiable neural dictionary (DND) \citep{pritzel2017neural}, with a cosine similarity look-up function. Lastly, $f_{decode}(z_{t+1})$ is implemented as a three layer multi-layer perceptron with $swish$ activation functions after each layer, except for the last, which utilizes a `sigmoid` activation function.

\subsection{Successor Similarity Algorithm}\label{SSA}

We use a modified form of the successor feature algorithm described in \citep{barreto2017successor}. The traditional formulation of policy learning using successor features consists of two functions, a reward function $w(s')$ and a successor function $\psi(s, s')$. These are updated using temporal difference learning as follows.

\begin{align}
    \delta_w = r_t - w(s') \\
    w(s')' = w(s') + \alpha_w \delta_w
\end{align}

Where $\alpha_w$ corresponds to the reward learning rate. 

\begin{align}
    \delta_{\psi} = s_t + \gamma \psi(s_{t+1}, a_{max}) - \psi(s_t, a_t) \\
    \psi(s_t, a_t)' = \psi(s_t, a_t). + \alpha_{\psi} \delta_{\psi}
\end{align}

Where $\alpha_{\psi}$ corresponds to the successor learning rate. $a_{max}$ corresponds to the action with the highest expected value, derived from the value function $Q(s, a) = \psi(s, a) \cdot w(s')^T$. This equation can also be used to derive a policy, where the $Q$ function can be used to derive a categorical distribution using a softmax function, i.e. $\pi(a | s)$.

While this works well in state spaces where each value in the state vector is independent of all others, it breaks down in cases where there is a mutual dependence. One example of this is over probabilistic state space representations, where a state vector $<s>$ would correspond to a belief state $<b>$ over state occupancies. In this scenario, there is no linear function of the state vector $s$ and a reward vector $w$ which would produce 1 when the agent is in the reward state and 0 in all other states. 

In order to address this issue, we replace the dot product with a cosine similarity function ($cos(A, B) = \frac{A \cdot B}{\norm{A} \norm{B}}$). This allows us to use probabilistic state representations from a generative temporal model as the state space, while bounding the reward and value functions between 0 and 1. We also replace the reward function update rule with one where $w(s')$ is set to $s'$ if a rewarding state is encountered. We additionally set $w(s')$ to a zero vector if the predicted reward was greater than 0.9, but a reward was not received in a given state . We note that this method is only viable in environments in which only a single state is rewarded at a time. Such a requirement however is not a limitation in goal-directed navigation tasks such as the ones performed here or often used in animal research. 

\subsection{Learning Hyperparameters}

\begin{table}[h!]
\begin{center}
\begin{tabular}{|p{4cm}|p{4cm}|}
 \hline
 \multicolumn{2}{|c|}{Generative Modeling Hyperparameters} \\
 \hline
 Parameter & Value \\
 \hline
 $z$ total size & 128 \\
 $z$ number distributions & 8 \\
 $s$ total size & 49 \\
 $s$ number distributions & 1 \\
 Learning rate & 5e-4 \\
 $h$ size & 256 \\
 $\beta_z$ & 0.001 \\
 $\beta_s$ & 0.001 \\
 Iterations & 5000 \\
 Batch size & 3 \\
 \hline
\end{tabular}
\end{center}
\caption{Hyperparameters used in WORLD and DSWM models in generative modeling experiments.}
\label{hyperparametersTable}
\end{table}

\begin{table}[h!]
\begin{center}
\begin{tabular}{|p{4cm}|p{4cm}|}
 \hline
 \multicolumn{2}{|c|}{Reinforcement Learning Hyperparameters} \\
 \hline
 Parameter & Value \\
 \hline
 $\gamma$ & 0.99 \\
 $\alpha$ & 0.1 \\
 Dyna rollout length & 5 \\
 Dyna rollout frequency & 0.2 \\
 $\tau$ & 0.001 \\
 \hline
\end{tabular}
\end{center}
\caption{Hyperparameters used for WORLD and DSWM models in reinforcement learning experiments.}
\label{hyperparametersTableRL}
\end{table}

\pagebreak

\subsection{Additional Figures}

\begin{figure}[h!]
\centering
\includegraphics[width=0.95\textwidth]{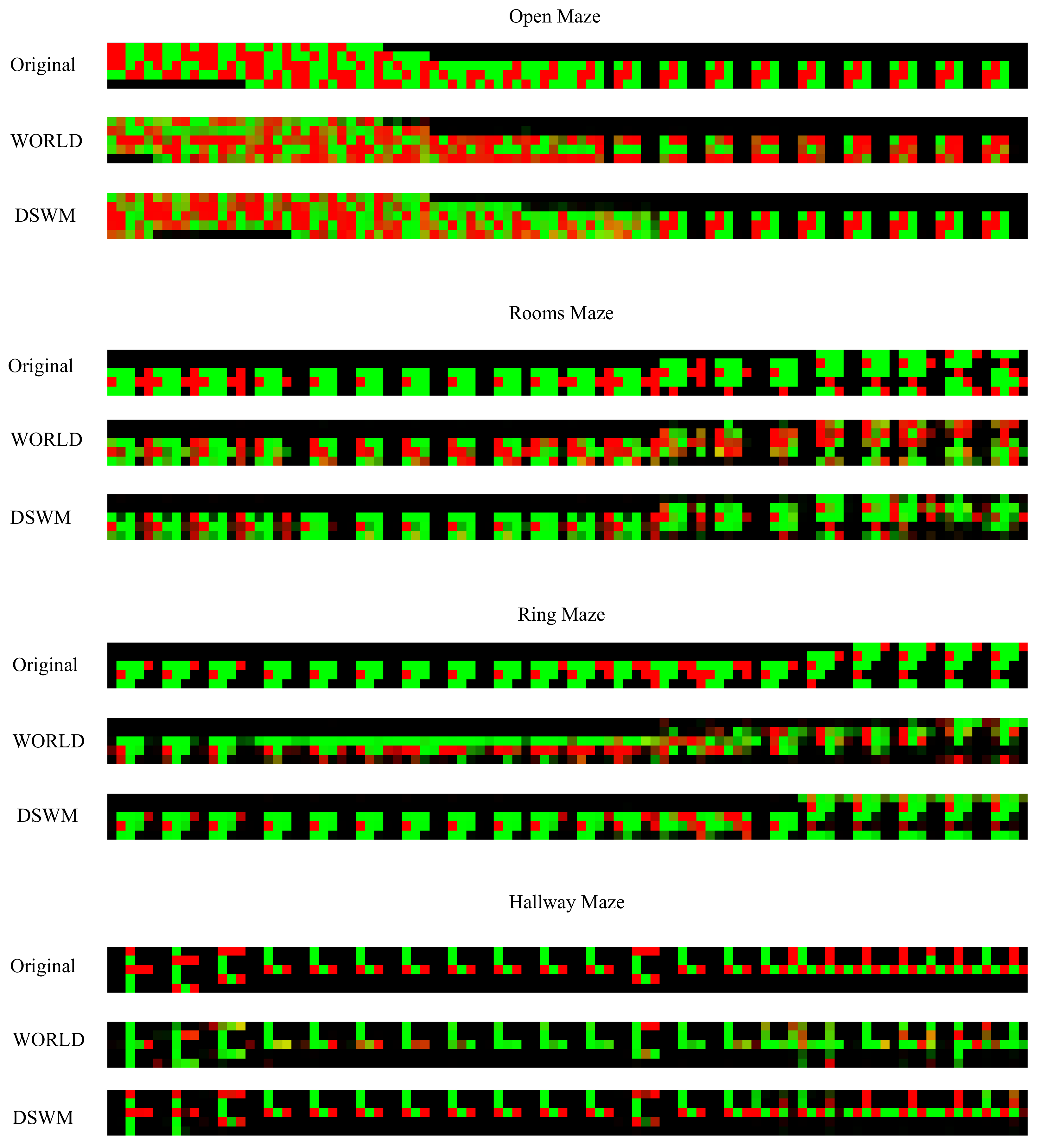}
\caption{Examples of reconstructed observations from rollouts of both World and DSWM models in all four environment topographies. Environments use pattern maps reserved for testing, and not seen during training. DSWM is able to better predict the true trajectory of future observations within all novel environments.}
\label{DSWM2DRolloutsFull}
\end{figure}

\begin{figure}[h!]
\centering
\includegraphics[width=1.0\textwidth]{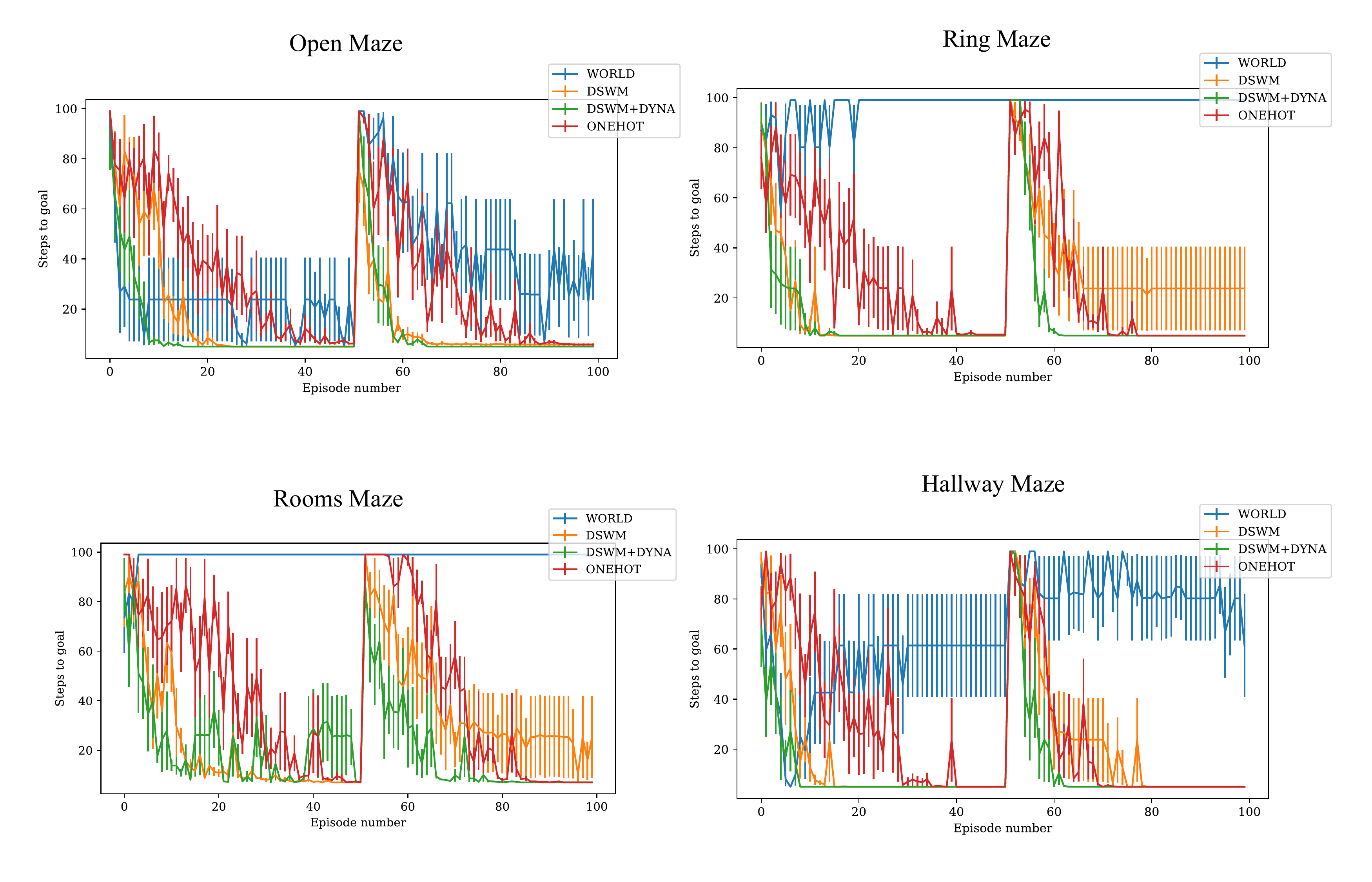}
\caption{Learning curves in goal-directed navigation task for each of the four unique environmental topographies. Each curve represents the average of five separate initialization seeds for the agent. Error bars represent standard error.}
\label{dswmRLResults}
\end{figure}

\begin{figure}[h!]
\centering
\includegraphics[width=0.5\textwidth]{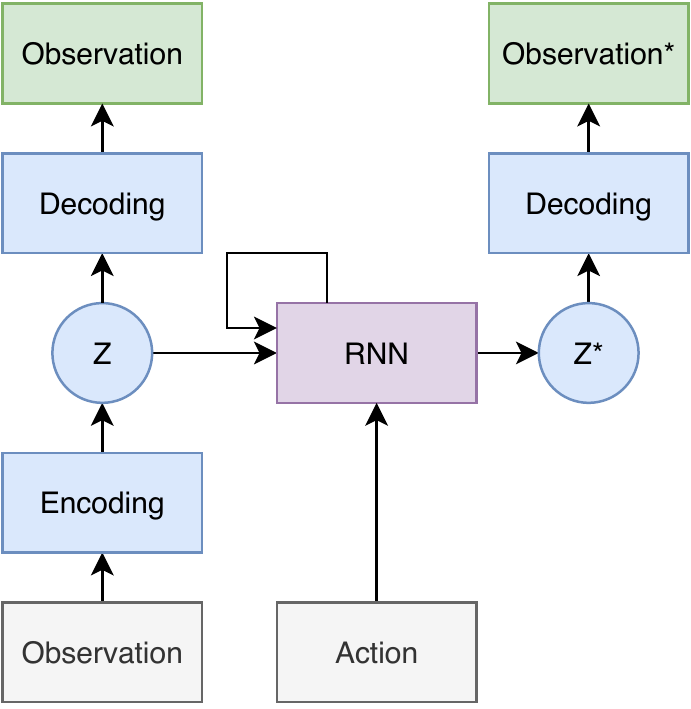}
\caption{Single Stream World Model diagram. Blue represents content information. Purple represents joint content and context information. White represents model inputs. Green represents model outputs. Nodes marked with a $*$ indicate information at the next time step of the simulation.}
\label{worldModelDiagram}
\end{figure}

\begin{figure}[h!]
\centering
\includegraphics[width=0.75\textwidth]{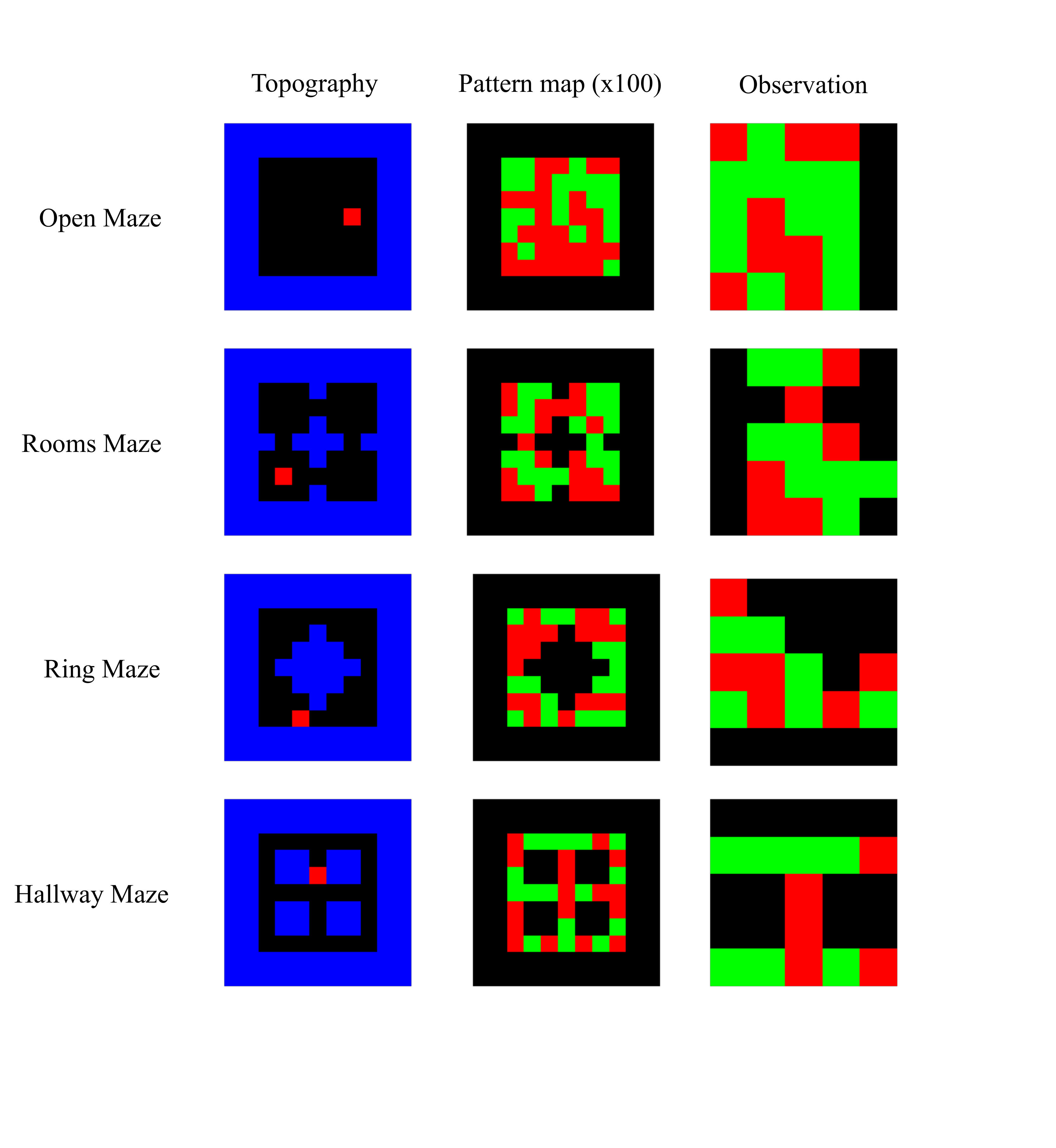}
\caption{Variable content environments with different topographies. Left: example environment topography. Blue corresponds to walls. Red corresponds to agent position. Middle: Randomly generated pattern image used to derive observations based on agent location. Right: Agent observations provide a 5x5 window around the agent position.}
\label{alloVariableAll}
\end{figure}

\end{document}